\documentclass[11pt]{article}
\usepackage{arxiv}
\usepackage{natbib}

\usepackage{amsmath,amssymb}
\usepackage{graphicx}
\usepackage{booktabs}
\usepackage{tabularx}
\usepackage{dcolumn}
\usepackage{microtype}
\usepackage[table]{xcolor}
\usepackage{float}
\usepackage{needspace}
\usepackage{caption}
\usepackage[hidelinks]{hyperref}
\usepackage{url}
\hypersetup{pdfauthor={Zelin Li and Caiwen Ding},pdftitle={Inference-Native Zeroth-Order Optimization for LLMs}}

\renewcommand{\arraystretch}{1.12}
\newcolumntype{d}[1]{D{.}{.}{#1}}
\newcolumntype{Y}{>{\raggedright\arraybackslash}X}

\newcommand{\system}{Infer-ZO}
\newcommand{\lozo}{LoZO}
\newcommand{\mezo}{MeZO}

\title{Inference-Native Zeroth-Order Optimization for LLMs}
\renewcommand{\shorttitle}{Inference-Native ZO for LLMs}
\author{
    Zelin Li\\
    University of Minnesota\\
    \texttt{li004593@umn.edu}
    \And
    Caiwen Ding\thanks{Corresponding author.}\\
    University of Minnesota\\
    \texttt{dingc@umn.edu}
}
\date{}

\begin{document}
\maketitle

\begin{abstract}
Zeroth-order (ZO) methods train large language models using only forward passes, yet common ZO execution paths perform substantial work beyond what the algorithm itself requires. To remove this extra execution overhead, we present \emph{Infer-ZO}, which separates the evaluations required by the algorithm from how they are executed, allowing them to reuse existing inference-engine optimizations. After removing inherited execution work, a complete Infer-ZO step on Qwen3-14B adds only 1.16\% wall-clock time over matched inference, bringing ZO execution close to its fundamental inference workload. With Infer-ZO, ZO evaluations run at near-inference cost with frozen base weights and can share GPU batches with ordinary inference requests. In co-serving experiments, foreground and background throughput remain within 1--3\% of the corresponding background-request baseline. Across 15 Qwen3, Llama, and OPT models, Infer-ZO achieves 2.10$\times$--9.94$\times$ end-to-end step speedups over released LoZO. Code is available at \url{https://github.com/playeriv65/zo-vllm}.
\end{abstract}

\section{Introduction}

Zeroth-order (ZO) methods train large language models (LLMs) using forward passes instead of backpropagation, avoiding the activation and gradient storage needed by a backward pass~\citep{malladi2023mezo,chen2025lozo}. In a two-sided step, the optimizer evaluates the loss at equal and opposite parameter perturbations, $L(\theta+\epsilon d)$ and $L(\theta-\epsilon d)$, and uses their difference to compute an update.

Since MeZO, ZO methods for LLMs have explored low-rank subspaces, memory offloading, on-device adaptation, evolution strategies, and modular algorithm toolkits~\citep{malladi2023mezo,chen2025lozo,wang2025zo2,gao2025mobizo,sarkar2025eggroll,dang2026zookit}. The algorithms have diversified, but their implementations remain strongly concentrated around the MeZO lineage. Among 22 post-MeZO methods with public implementations, 18 (81.8\%) directly or transitively descend from MeZO, including 16 (72.7\%) that retain or modify its training code.

This lineage preserves a familiar execution path even as the algorithms change. A typical MeZO-style two-sided step performs four dense weight writes: write the weights with the positive perturbation, write them with the negative perturbation, restore the unperturbed weights, and write the update. The resulting cost is tied to this dense-mutation path even when an algorithm admits a more structured representation. More broadly, coupling the algorithm to its execution path makes execution-specific inefficiencies harder to isolate. This leads to a more basic question: \emph{which operations are required by the ZO algorithm, and which are inherited from its execution path?}

We answer this question by following the data flow of a ZO step through Direction, Query, Estimator, and Update. This separates the evaluations requested by the algorithm from the backend that carries them out. We call ZO execution \emph{inference-native} when its function evaluations can be issued as inference requests. An inference runtime can then execute, batch, and schedule these evaluations through its normal request path.

\begin{figure}[t]
    \centering
    \includegraphics[width=0.98\linewidth]{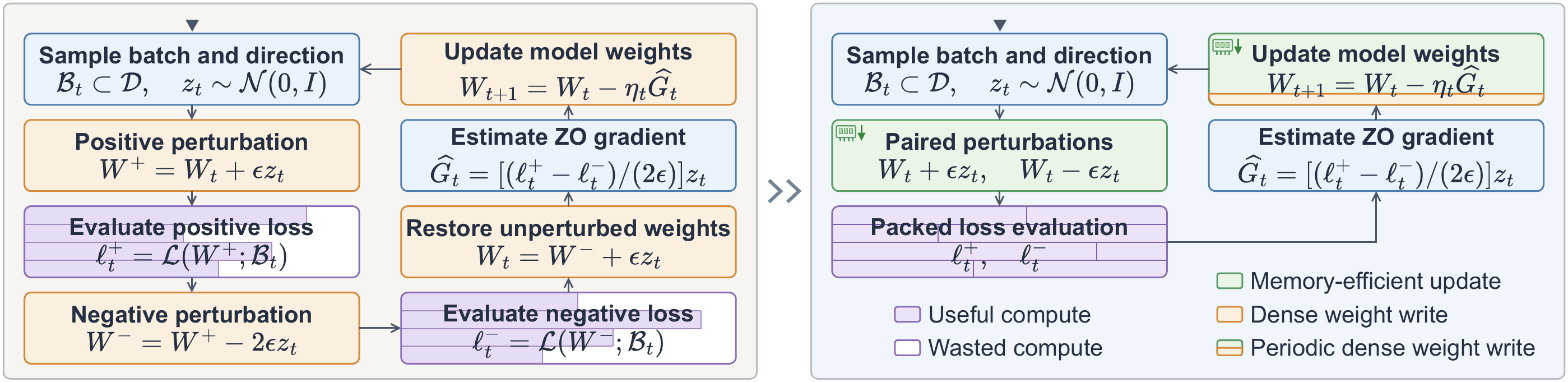}
    \caption{Conventional and inference-native ZO execution. The conventional path materializes the positive perturbation, negative perturbation, restoration, and optimizer update as separate dense weight writes around the two loss evaluations. The Infer-ZO path evaluates paired probes efficiently, combines operations in the conventional loop, and avoids intermediate dense weight writes, reducing memory traffic.}
    \label{fig:overview}
\end{figure}

Infer-ZO realizes this boundary with low-rank directions. Their compact factors avoid repeated dense weight writes; keeping the base weights frozen also lets ZO and serving requests share one inference engine. Figure~\ref{fig:overview} contrasts this path with conventional dense-mutation ZO execution.

\Needspace{14\baselineskip}
Our contributions are:
\begin{enumerate}
    \item \textbf{We decouple ZO algorithms from execution.} Direction generates perturbations, Query packages them into the required function-evaluation requests, Estimator reduces the returned observations, and Update determines the parameter change. Different ZO methods can therefore reuse the same execution backend.
    \item \textbf{We remove execution work not required by the algorithm.} Low-rank Direction lets the backend keep parameter changes in compact factors instead of repeatedly rewriting dense weights. This brings ZO execution close to the cost of the function evaluations that the algorithm itself requires.
    \item \textbf{We enable ZO training and inference to share an inference engine.} With frozen base weights, ZO evaluations attach request-specific low-rank factors while ordinary serving requests use the base weights directly. Both can therefore share GPU batches and the same inference engine.
\end{enumerate}

Across 15 Qwen3, Llama, and OPT models, \system{} achieves 2.10$\times$--9.94$\times$ step speedups over released LoZO, while a Qwen3-14B step adds only 1.16\% over matched inference. Learning follows released LoZO, and co-serving changes foreground and background throughput by only 1--3\%.

\section{ZO Algorithms and Execution Requirements}
\label{sec:layered-zo}

To change execution without changing the algorithm, we first describe a ZO step independently of any backend.

\subsection{Algorithmic Components}
\label{sec:algorithm-setup}

At step $t$, let $\theta_t$ denote the selected parameters of a model $f_{\theta_t}$. A two-sided ZO step samples direction $d_t$ and evaluates loss $L$ on the same input $\xi_t$ at $\theta_t\pm\epsilon d_t$:
\begin{equation}
    c_t = \frac{L(\theta_t+\epsilon d_t;\xi_t)-L(\theta_t-\epsilon d_t;\xi_t)}{2\epsilon},
    \qquad \Delta\theta_t=-\eta_t c_t d_t.
    \label{eq:antithetic-example}
\end{equation}
Here $\epsilon$ is the perturbation scale, $\eta_t$ is the learning rate, and $c_t$ estimates the directional derivative of the loss along $d_t$.

A ZO step has four responsibilities. \emph{Direction} produces one or more search directions and maintains the state needed to reconstruct them, such as seeds, low-rank factors, or sparse support. \emph{Query} uses those directions to form the function-evaluation requests handed to the backend, including the requested parameter states, inputs, and dependencies. The backend executes the requests and returns observations such as losses or rewards. \emph{Estimator} converts those observations into a gradient estimate or search signal. Unlike first-order training, where backpropagation largely fixes gradient construction, ZO methods can differ in how observations are converted into an estimate. \emph{Update} combines that estimate with optimizer state to determine the parameter change and the next optimizer state, including SGD-, momentum-, Adam-, or variance-reduction-style rules.

These four roles can vary separately across ZO methods. A low-rank direction $UV^T$, for example, is a Direction choice, while a one-sided evaluation, an antithetic pair, a center evaluation with perturbations, or a population of evaluations is a Query choice. Estimator determines how the returned observations become an estimate, and Update determines how that estimate and optimizer state change the parameters.

This decomposition also reflects how LLM ZO methods differ in practice. Among 30 post-MeZO parameter-space ZO methods, 25 (83.3\%) modify no more than two of the four components.

\noindent
\begin{minipage}[t]{0.31\linewidth}
\vspace{0pt}
Most algorithmic changes are therefore local to part of the ZO step even when implementations reuse much of the same training machinery. ZOOkit~\citep{dang2026zookit} organizes ZO methods along five configuration axes; our decomposition instead tracks what the backend must preserve. Query hands the backend a set of evaluation requests, while the backend decides how to execute them. This lets execution change without redefining the ZO method.
\end{minipage}\hfill
\begin{minipage}[t]{0.66\linewidth}
\vspace{0pt}
    \centering
    \includegraphics[width=\linewidth]{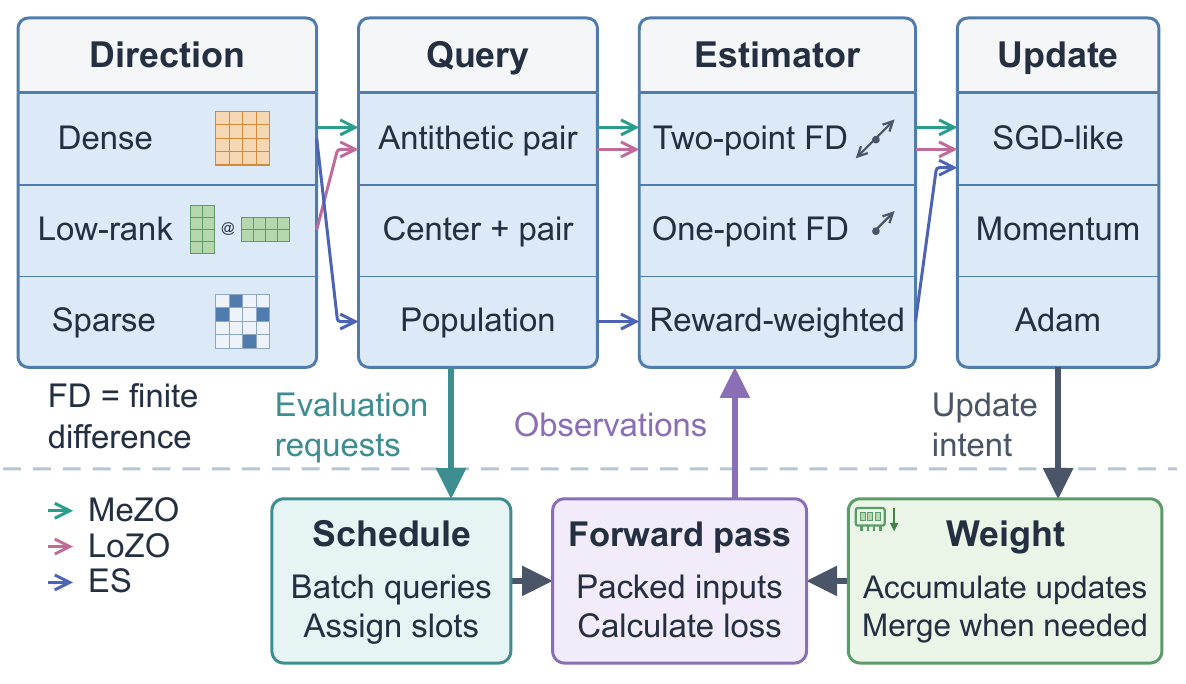}
    \captionof{figure}{The four algorithm components can be recombined across ZO methods. The evaluation-request boundary keeps the algorithm separate from backend execution.}
    \label{fig:query-boundary}
\end{minipage}

\subsection{Separating Algorithm and Execution}
\label{sec:composition}

Once Query has formed the evaluation requests for a ZO step, those requests form the contract with the backend. The backend must evaluate the requested parameter states on the requested inputs, respect dependencies among requests, and return the required observations. It does not need to reproduce the physical operation sequence of a reference trainer. Independent evaluation requests may be batched as long as the backend preserves the specified parameter states and dependencies.

The distinction between what a request asks for and how the backend executes it is already familiar in ML systems. Prefix caching reuses previously computed KV state for requests with a shared prompt prefix, while FlashAttention changes the physical execution of exact attention without changing its mathematical result~\citep{vllm2026prefixcaching,dao2022flashattention}. Variable-length token-scoring requests likewise require scores for the actual tokens, not computation on padding positions, so a backend can pack ragged inputs instead.

The same freedom applies to parameter state. For one weight matrix, write $W_t=W_{\mathrm b}+S_t$, where $W_t$ is the parameter value requested at step $t$, $W_{\mathrm b}$ the stored base matrix, and $S_t$ the accumulated parameter change. An evaluation request may require the parameter state $W_t$ without requiring the backend to materialize $W_t$ as a new dense matrix; the backend may instead compute with $W_{\mathrm b}$ and $S_t$ separately.

For a low-rank Direction, the requested parameter state has the same additive form already used by LoRA serving. For a weight matrix $W\in\mathbb R^{m\times n}$, a low-rank direction has the form $UV^T$, with $U\in\mathbb R^{m\times r}$ and $V\in\mathbb R^{n\times r}$. LoRA inference can apply this factor on top of the base weight without first materializing the corresponding dense weight~\citep{hu2022lora}. The low-rank form is also preserved under linear combinations when directions share the same right factor. For coefficients $a_j$,
\[
    \widehat G=\sum_{j=1}^{N}a_jU_jV^T
              =\left(\sum_{j=1}^{N}a_jU_j\right)V^T.
\]
The aggregate has rank at most $r$, independent of $N$. The same algebra lets updates across consecutive steps sharing $V$ accumulate by adding their coefficient-weighted left factors through a complete ZO loop.

For methods defined with dense Gaussian directions, the decomposition also lets us change Direction while keeping Query, Estimator, and Update fixed. A Gaussian-product family gives a precise distributional connection to the original Direction. Let $Z_r=\frac{1}{\sqrt r}UV^T$, where $U\in\mathbb R^{m\times r}$ and $V\in\mathbb R^{n\times r}$ have independent standard-normal entries. As $r$ grows, any fixed finite set of entries of $Z_r$ converges jointly to independent standard Gaussians~\citep{sarkar2025eggroll}. Thus rank gives a controlled route from a compact factorized Direction toward the dense-Gaussian distribution. At finite rank, $Z_r$ is still a different Direction rather than an execution-only rewrite.

\section{ZO Training as an Inference Workload}
\label{sec:inference-workload}

Low-rank directions let the requested perturbations and accumulated updates remain as additive low-rank factors. We now show how a complete ZO step runs with the base weights and these factors kept separate, using vLLM as the reference backend~\citep{kwon2023vllm}.

Throughout the paper, $t$ remains the global optimization-step index and $T$ the total step budget. To make the sequence concrete, we first follow the initial fixed-$V$ stage. Its local steps are $s=0,\ldots,\nu-1$, with global step $t=s$. During this stage, the right factor $V^{(0)}$ is fixed and a fresh left factor $U_s^{(0)}$ is sampled at each step. If $W$ denotes the unperturbed weight used by the current evaluation, the positive and negative perturbations act on a layer input $x$ as
\[
    W^{\pm}x=Wx\pm\frac{\epsilon}{\sqrt r}U_s^{(0)}\bigl((V^{(0)})^Tx\bigr).
\]
LoRA inference already computes a base weight plus additive low-rank factors, so the backend can apply both perturbations and accumulated updates through the same low-rank forward path.

\subsection{Accumulation within a Fixed Subspace}
\label{sec:fixed-subspace-accumulation}

\begin{figure}[H]
    \centering
    \includegraphics[width=0.98\linewidth]{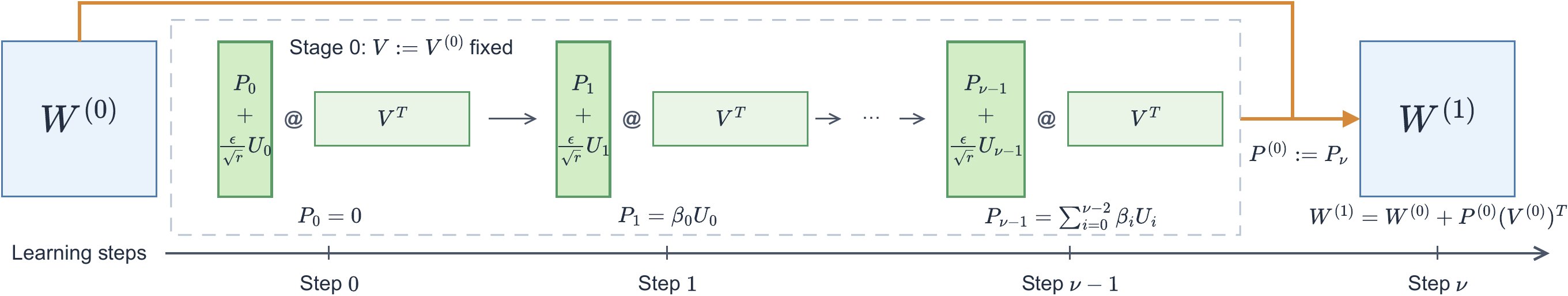}
    \caption{Accumulation within one fixed subspace. $V^{(0)}$ remains fixed for $\nu$ local steps while $P_s^{(0)}$ accumulates the coefficient-weighted left factors. After the stage, the completed update $P^{(0)}(V^{(0)})^T$ may be folded into a writable base before the next subspace begins.}
    \label{fig:fixed-subspace-accumulation}
\end{figure}

Executing the requested evaluations is only part of the loop: the next evaluation must also include the updates accumulated so far. Within the initial stage, every update shares the same right factor $V^{(0)}$. We therefore accumulate only the left factor. Let $P_s^{(0)}$ be the update accumulated after the first $s$ local steps, with $P_0^{(0)}=0$:
\[
    P_s^{(0)}=\sum_{i=0}^{s-1}\beta_i^{(0)}U_i^{(0)},\qquad
    P_{s+1}^{(0)}=P_s^{(0)}+\beta_s^{(0)}U_s^{(0)}.
\]
For the SGD-like rule in Equation~\ref{eq:antithetic-example}, $\beta_s^{(0)}$ contains the learning rate, finite-difference coefficient, and $1/\sqrt r$ normalization for global step $t=s$. With $W_{\mathrm b}$ as the stored base, the two evaluations at local step $s$ use
\[
    W^{\pm}=W_{\mathrm b}+\left(P_s^{(0)}\pm\frac{\epsilon}{\sqrt r}U_s^{(0)}\right)(V^{(0)})^T.
\]
After the step, $P_{s+1}^{(0)}$ carries the new update into the next evaluation. Figure~\ref{fig:fixed-subspace-accumulation} shows this accumulation over the initial fixed-$V$ stage.

\subsection{LoRA Bank across Subspaces}
\label{sec:lora-bank}

After $\nu$ steps, the initial stage leaves the completed pair $(P^{(0)},V^{(0)})$, where $P^{(0)}:=P_{\nu}^{(0)}$. Refreshing $V$ starts the next stage with a fresh accumulator $P_0^{(1)}=0$. Repeating this process produces completed pairs $(P^{(k)},V^{(k)})$, $k=0,1,\ldots$, with $P^{(k)}:=P_{\nu}^{(k)}$. If a completed update is not merged into the base weights, we retain its pair in the LoRA bank.

During stage $k$, if all earlier completed stages remain outside the base weights, the two evaluations use
\begin{equation}
    W^{\pm}=W_{\mathrm b}+\sum_{j<k}P^{(j)}(V^{(j)})^T
                    +\left(P_s^{(k)}\pm\frac{\epsilon}{\sqrt r}U_s^{(k)}\right)(V^{(k)})^T.
    \label{eq:combined-probe}
\end{equation}
The first sum contains updates from completed stages; $P_s^{(k)}$ contains only the updates accumulated within the current stage. Retaining the completed pairs delays dense writes across subspaces. In our vLLM backend, these pairs occupy GPU LoRA adapter slots.

If the base weights are writable, as in ordinary standalone training, completed bank entries may be merged whenever convenient between evaluations. Merge timing is an execution choice; subsequent evaluations continue from the updated base weights.

\begin{figure}[H]
    \centering
    \includegraphics[width=0.98\linewidth]{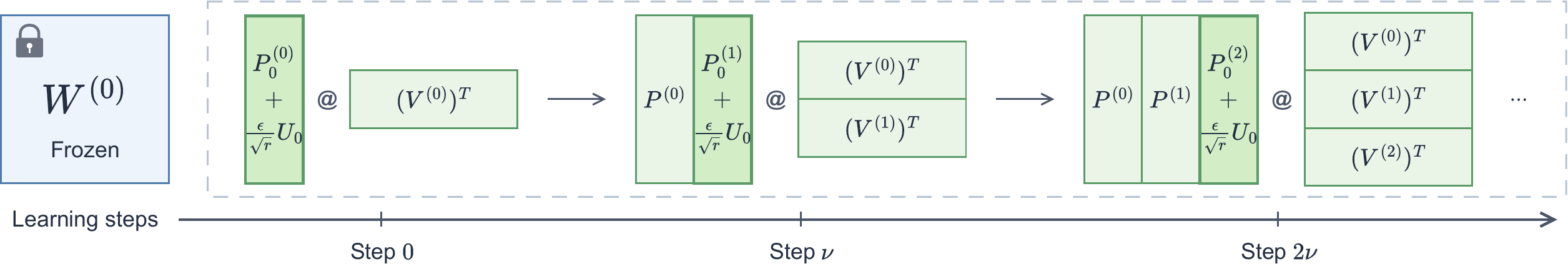}
    \caption{Frozen base weights across subspace refreshes. The LoRA bank keeps completed pairs $(P^{(j)},V^{(j)})$ as new subspaces are introduced, while $W_0$ remains unchanged.}
    \label{fig:state-mechanism}
\end{figure}

If the base weights are to remain frozen, the bank can instead retain every stage's factors from the start. A run of $T$ total optimization steps with refresh interval $\nu$ has at most $\lceil T/\nu\rceil$ stages. Each stage contributes rank $r$, so the required total inner dimension is $r\lceil T/\nu\rceil$. With $r=2$, $T=20{,}000$, and $\nu=500$, this gives an inner dimension of 80. LoZO reports comparable SST-2 accuracy at $\nu=500$ relative to smaller refresh intervals~\citep{chen2025lozo}. This capacity is known before the run, allowing the base weights to remain at their initial value $W_0$ throughout training. A fixed-capacity bank can instead use \emph{cyclic subspace reuse}: once the bank is full, training cycles back through the stored right factors $V^{(j)}$ and continues accumulating their corresponding left factors $P^{(j)}$, keeping the total inner dimension bounded by the bank capacity. vLLM fixes the maximum LoRA rank when the engine is created, so the bank capacity is configured up front. Figure~\ref{fig:state-mechanism} shows the frozen-base case.

\subsection{Sharing the Inference Engine}
\label{sec:serving-design}

With $W_0$ frozen, each ZO evaluation can attach its own low-rank factors without changing the shared base model. The optimizer can therefore submit the evaluation as an inference request. Independent requests can be batched; dependent requests wait for the returned loss or reward before the optimizer advances its state.

\begin{figure}[H]
    \centering
    \includegraphics[width=0.98\linewidth]{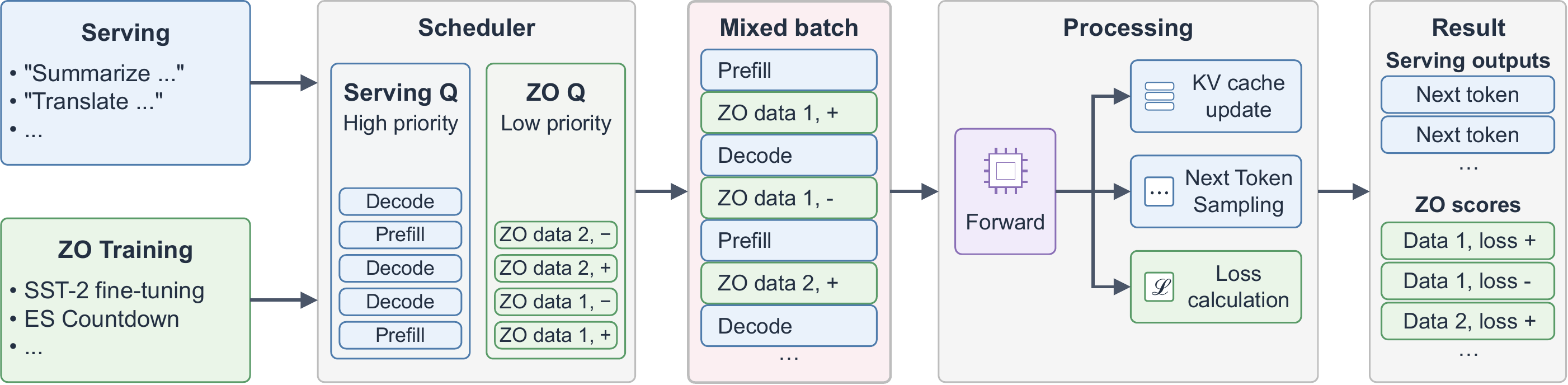}
    \caption{Co-serving through one inference engine. Serving and ZO requests enter separate scheduler queues and may be interleaved in the same GPU batch. They share the forward path, after which serving requests continue to KV-cache update and next-token sampling while ZO requests return loss values to the optimizer.}
    \label{fig:architecture}
\end{figure}

Foreground serving requests use $W_0$. Inference requests for ZO evaluations attach the completed bank pairs together with the current-stage factors and perturbation from Equation~\ref{eq:combined-probe}. The engine can place both request types in one GPU batch without rewriting $W_0$. The serving scheduler decides when the background evaluations are admitted (Figure~\ref{fig:architecture}).

\section{Evaluation}
\label{sec:evaluation}

\paragraph{Experimental setup.}\label{sec:experimental-setup}
Correctness experiments use SST-2, BoolQ, and CB~\citep{socher2013recursive,wang2018glue,wang2019superglue}. We use the released LoZO implementation~\citep{chen2025lozo} as the baseline for learning and end-to-end timing. Systems experiments cover 15 Qwen3, Llama, and OPT models from 350M to 32B parameters at batch 16. Unless stated otherwise, LoZO-style runs use rank $r=2$ and refresh interval $\nu=50$. Matched inference removes direction generation, perturbation preparation and application, and the final update while keeping the same request path; co-serving compares the ZO background with the corresponding LoRA-prefill background under the same foreground load.

\subsection{Correctness and Training Results}
\label{sec:correctness-eval}

We compare against the released LoZO implementation~\citep{chen2025lozo} on SST-2, BoolQ, and CB~\citep{socher2013recursive,wang2018glue,wang2019superglue}. Over 60 joined steps on OPT-1.3B, projected-gradient signs agree on 60/60 steps on SST-2, 58/60 on BoolQ, and 60/60 on CB, with zero seed or refresh mismatches. On OPT-13B SST-2, signs agree on 59/60 steps and the projected-gradient Pearson correlation is 0.99896. The complete generation--reward--update loop also trains on the ES-at-Scale Countdown workload.

\begin{figure}[H]
    \centering
    \includegraphics[width=0.98\linewidth]{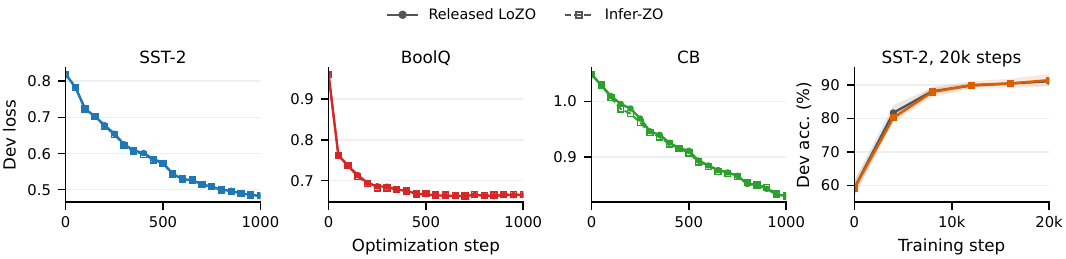}
    \caption{Learning correspondence with released LoZO: 1,000-step development-loss trajectories on SST-2, BoolQ, and CB, and 20k-step SST-2 development accuracy (mean $\pm$ sample standard deviation over three seeds).}
    \label{fig:correctness-training}
\end{figure}

We then extend the scoring comparison to strict 1,000-step learning curves. Across SST-2, BoolQ, and CB, the maximum dev-loss gap is 1.04\%, 1.30\%, and 4.17\% of the corresponding reference loss drop (Figure~\ref{fig:correctness-training}, left three panels). Three-seed endpoints at step 1,000 remain close to released LoZO (Table~\ref{tab:learning-alignment}). Pooling all 45 dev-loss checkpoints across the three tasks gives Pearson correlation 0.9946 and linear-fit $R^2=0.9893$.

Over 20k OPT-1.3B steps on SST-2, the three-seed development-accuracy trajectories remain close (Figure~\ref{fig:correctness-training}, right panel). At the final checkpoint, mean development accuracy is 91.07\% for released LoZO and 91.40\% for \system{}; the largest per-seed gap across recorded checkpoints is 0.024.

\begin{table}[H]
    \centering
    \small
    \caption{Three-seed learning comparison with the released LoZO implementation~\citep{chen2025lozo}, used as our learning baseline on OPT-1.3B. Entries are mean $\pm$ sample standard deviation at step 1,000.}
    \label{tab:learning-alignment}
    \begin{tabular}{lrrrr}
        \toprule
        & \multicolumn{2}{c}{Final dev. loss} & \multicolumn{2}{c}{Final dev. accuracy} \\
        \cmidrule(lr){2-3}\cmidrule(lr){4-5}
        Task & Released LoZO & \system{} & Released LoZO & \system{} \\
        \midrule
        SST-2 & $0.4859\pm0.0179$ & $0.4959\pm0.0116$ & $0.779\pm0.014$ & $0.768\pm0.004$ \\
        BoolQ & $0.6522\pm0.0134$ & $0.6546\pm0.0206$ & $0.614\pm0.026$ & $0.614\pm0.024$ \\
        CB & $0.8732\pm0.0391$ & $0.8421\pm0.0212$ & $0.527\pm0.049$ & $0.593\pm0.035$ \\
        \bottomrule
    \end{tabular}
\end{table}

\subsection{Removing Inherited Execution Work}
\label{sec:source-attribution}
\label{sec:external-state-cost}

\begin{figure}[H]
    \centering
    \includegraphics[width=0.99\linewidth]{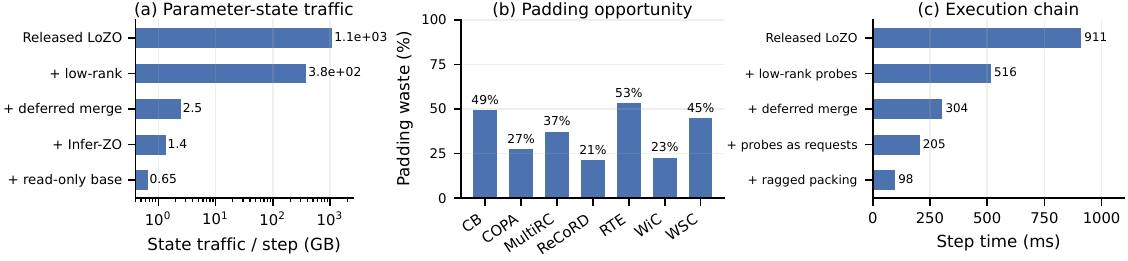}
    \caption{Removing inherited execution work on Qwen3-14B: (a) parameter-state traffic, (b) padding waste, and (c) step time through the execution chain. The probes-as-requests stage keeps the same padded batches; ragged packing is enabled next.}
    \label{fig:execution-peeling}
\end{figure}

With numerical and learning correspondence established, we next ask which parts of the inherited execution path are not required by the queried evaluations themselves. Figure~\ref{fig:execution-peeling}(a) shows the parameter-state traffic on Qwen3-14B. The released dense path moves 1,092.9~GB per step; low-rank perturbation reduces this to 384.0~GB, deferred merge to 2.47~GB, and \system{} to 1.35~GB. With a read-only base, the remaining parameter-state traffic is 0.65~GB per step.

The backend can also avoid padded-token computation. Figure~\ref{fig:execution-peeling}(b) shows the padding fraction for Qwen-tokenized SuperGLUE requests at batch 16: 21--53\% of padded tokens are empty work across the seven tasks. With the token layout fixed, packed PyTorch and vLLM forwards differ by 0.1\% on OPT-13B SST-2 and BoolQ.

Figure~\ref{fig:execution-peeling}(c) follows the complete Qwen3-14B chain: 911.01~ms for dense perturbation and dense update, 516.33~ms after low-rank perturbation, 303.77~ms after deferred merge, 205.09~ms when the probes are issued as inference requests on the same padded batches, and 97.62~ms after enabling ragged packing.

\subsection{Cost Relative to Matched Inference}
\label{sec:matched-inference-eval}
\label{sec:cost-prediction}

The peeling experiment suggests that once dense mutations are removed, little cost remains beyond the inference work itself. We therefore compare complete ZO steps against matched inference, which removes direction generation, perturbation preparation and application, and the final update while keeping the same request path. On Qwen3-14B, \system{} takes 92.99~ms versus 91.93~ms for matched inference, or 1.16\% $\pm$ 0.40 percentage points of overhead.

\begin{figure*}[t]
    \centering
    \includegraphics[width=0.95\textwidth]{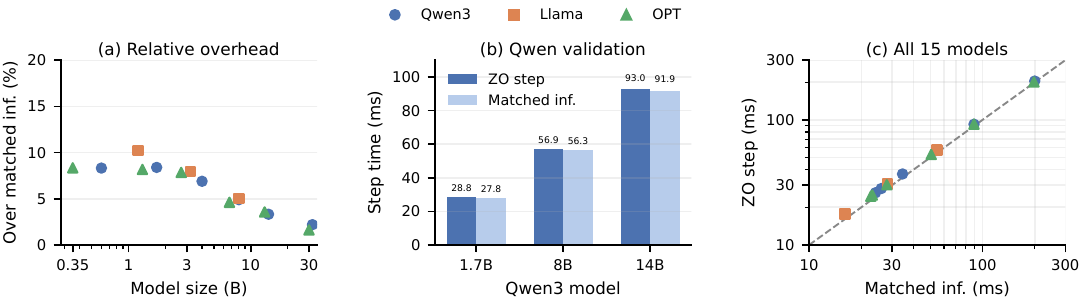}
    \caption{\system{} versus matched inference: (a) relative overhead across 15 models, (b) direct Qwen3 matched-inference validation at selected scales, and (c) the all-model identity-line view; the dashed line marks parity.}
    \label{fig:matched-inference-scaling}
\end{figure*}

Across the 15-model sweep from 350M to 32B parameters, the corresponding overhead ranges from 1.63\% to 10.23\% of matched inference and is at most 5.03\% from 6.7B upward (Figure~\ref{fig:matched-inference-scaling}(a)). On direct Qwen3 validation runs, the paired step times are 28.79/27.83, 56.91/56.31, and 92.99/91.93~ms for 1.7B, 8B, and 14B, respectively (Figure~\ref{fig:matched-inference-scaling}(b)); the full 15-model comparison remains close to the identity line (Figure~\ref{fig:matched-inference-scaling}(c)).

\subsection{End-to-End Speedup over Released LoZO}
\label{sec:released-lozo-speedup}

\begin{figure}[H]
    \centering
    \includegraphics[width=0.80\linewidth]{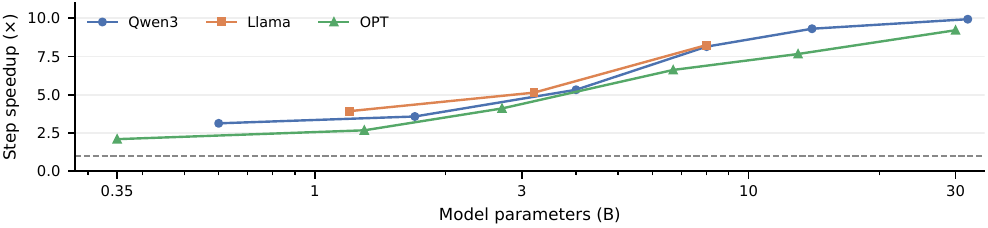}
    \caption{End-to-end SST-2 step speedup over released LoZO across 15 Qwen3, Llama, and OPT models; the dashed line marks 1$\times$.}
    \label{fig:released-lozo-speedup}
\end{figure}

The matched-inference comparison measures closeness to the inference workload. We now compare against the released LoZO implementation to quantify the practical end-to-end benefit. On SST-2, \system{} is 2.10$\times$--9.94$\times$ faster per optimization step than released LoZO across 15 Qwen3, Llama, and OPT models (Figure~\ref{fig:released-lozo-speedup}). Speedup grows with scale across all three families: Qwen3-32B reaches 9.94$\times$, Llama-3.1-8B reaches 8.25$\times$, and OPT-30B reaches 9.22$\times$. The magnitude is workload-dependent: on BoolQ, Qwen3-14B reaches 3.50$\times$, while OPT-1.3B and OPT-13B reach 2.96$\times$ and 2.95$\times$.

\subsection{Co-Serving with Ordinary Inference}
\label{sec:mixed-batch-eval}

Keeping the base weights frozen also lets ZO evaluations and serving requests share the same engine. On Qwen3-8B with batch-invariant execution~\citep{vllm2026batchinvariance}, 31 mixed batches across batch sizes 4, 8, and 16 reproduce the standalone serving outputs, $L_+$, $L_-$, and directional finite difference exactly.

\begin{figure}[H]
    \centering
    \includegraphics[width=0.88\linewidth]{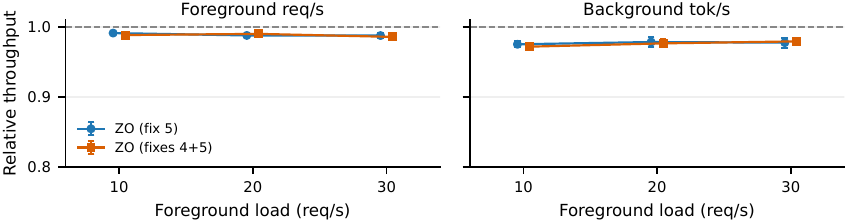}
    \caption{Co-serving throughput normalized by the corresponding LoRA-prefill background request stream at the same foreground load. The dashed line marks parity; each point averages four windows.}
    \label{fig:coserving-throughput}
\end{figure}

We then compare shared-engine throughput when the background workload is a LoRA-prefill stream or the same stream plus ZO updates. Under foreground loads of 10, 20, and 30 requests/s, the two ZO arms retain 98.6--99.1\% of foreground request throughput and 97.2--97.9\% of background scheduled-token throughput relative to the corresponding LoRA-prefill background baseline (Figure~\ref{fig:coserving-throughput}).

\section{Related Work}

LLM ZO methods modify different, largely orthogonal parts of the optimization loop. LoZO uses reusable low-rank subspaces for Direction, ZoAR reuses previous evaluations in Query and Estimator, AdaMeZO introduces an Adam-style Update rule, and EGGROLL uses low-rank Gaussian-product perturbations for large-population evolution strategies~\citep{chen2025lozo,qiu2025zoar,cai2026adamezo,sarkar2025eggroll}. Despite this algorithmic diversity, the dominant implementation lineage remains a tightly coupled MeZO-style trainer loop, carrying a common execution path across otherwise different algorithms. Infer-ZO makes these component boundaries explicit and separates the resulting evaluation requests from backend execution.

Prior systems address particular ZO execution settings. ZO2 streams model blocks between CPU and GPU and overlaps parameter transfers with the longer dual-forward computation~\citep{wang2025zo2}. Infer-ZO changes the underlying execution path before offloading by keeping perturbations and accumulated state over a frozen base. MobiZO runs multi-query ZO through ExecuTorch~\citep{gao2025mobizo}, with perturbation and update logic encoded in an MP-LoRA model graph. Infer-ZO instead decouples the algorithmic components from one another, separates the algorithm from backend execution, and studies how closely ZO execution can approach matched inference while sharing the engine with ordinary inference requests.

LoRA keeps a shared base model fixed while representing model changes with low-rank adapters~\citep{hu2022lora}. Multi-LoRA serving systems such as Punica efficiently batch requests that share the same frozen base while using different adapters~\citep{chen2023punica}. Infer-ZO uses this execution form when requested ZO parameter states admit a low-rank representation over a shared frozen base. Unlike conventional multi-LoRA serving, these low-rank states are generated and updated along an optimization trajectory rather than selected from a fixed set of task adapters.

Once ZO evaluations are expressed as inference requests, online/offline serving mechanisms become directly relevant. HyGen co-locates latency-sensitive online requests with offline inference~\citep{sun2025hygen}; DeltaServe, FlexLLM, and LLMStation similarly study accelerator sharing between serving and tuning or training~\citep{chen2026deltaserve,oliaro2026flexllm,he2025llmstation}. Our co-serving experiments place ZO requests in this background role alongside ordinary inference.

\section{Discussion}

Sharing an execution engine between inference and ZO suggests a different way to think about adaptation during serving. Live requests could themselves become optimization queries: a small fraction could be evaluated under perturbed parameter states, while subsequent user feedback, task outcomes, or automated rewards provide the scalar observations needed by ZO. These observations could accumulate across requests and be applied through deferred updates, allowing data collection, evaluation, and optimization to occur within the serving system itself. This possibility suggests a path toward continual adaptation driven directly by serving traffic. Because ZO needs only scalar observations, the feedback source need not expose gradients or retain the serving computation graph.

\section{Conclusion}

\system{} separates ZO algorithmic components from backend execution by expressing function evaluations as inference requests. With compatible low-rank directions, this removes repeated dense weight writes, approaches matched inference, and allows ZO and ordinary inference requests to share an engine. This reframes ZO training as inference-compatible model evaluations plus lightweight state management.

\section{Limitations}

Our low-rank realization can approximate dense-Gaussian-like directions through CLT-style constructions, although finite-rank perturbations remain structured rather than exactly dense Gaussian. Our co-serving study focuses on execution compatibility rather than serving-policy design; integrating \system{} with schedulers such as HyGen~\citep{sun2025hygen} is a natural next step.

\clearpage
\section*{AI Use Statement}
We used generative AI tools to assist with manuscript drafting, editing, and organization; LaTeX formatting; code generation and debugging for implementation and experimental scripts; quantitative data analysis; literature search and summarization; and discussions of experimental design and result presentation. All AI-assisted outputs, including code, analyses, references, and manuscript text, were reviewed and verified by the authors. The authors take full responsibility for the final content of this work, including all claims, results, and artifacts produced with the assistance of generative AI.

\section*{Reproducibility Statement}
We organize the reproducibility material around the claims in the main paper. Section~\ref{sec:experimental-setup} specifies the evaluated models and workloads, baselines and controls, metrics, and repetition policies. Appendix~\ref{app:experimental-scope} records optimization configurations and dataset scope, and Appendix~\ref{app:environment} summarizes the recorded hardware and execution environments. Appendix~\ref{app:correctness} provides the numerical-alignment and learning-comparison details, Appendix~\ref{app:es-evidence} records the generation-based Query experiment, and Appendix~\ref{app:timing-accounting} gives the timing definitions and standalone measurements used in the reported results. Appendix~\ref{app:implementation-scope} maps the four components to implementation objects and shows how the backend stores parameter state; the Gaussian-product statement is proved in Appendix~\ref{app:clt-proof}.

\bibliographystyle{plainnat}
\bibliography{references}

\clearpage
\appendix
\raggedbottom
\section*{Appendix}
\section{Literature Audit Under the Four-Component ZO Decomposition}\label{app:taxonomy}

We audit 30 named post-MeZO parameter-space ZO methods, using MeZO as the comparison baseline. The corpus includes language-model post-training methods with an algorithmic ZO mechanism that maps to the four components and is described in enough detail to identify which components it changes; method families are counted once. We exclude execution-only systems and explicit first-order/ZO hybrids. The audit cutoff is September 17, 2026.

For each method, we record whether it modifies Direction (D), Query (Q), Estimator (E), or Update (U) relative to MeZO and trace code lineage when a public implementation is available. Eleven methods modify exactly one component, 14 modify two, and 5 modify three; none modifies all four. Thus 25/30 (83.3\%) modify no more than two components. Direction is modified by 20/30 methods (66.7\%), Query by 13/30 (43.3\%), Estimator by 13/30 (43.3\%), and Update by 8/30 (26.7\%). Among 22 methods with public code, 18 (81.8\%) directly or transitively descend from the MeZO training codebase and 16 (72.7\%) retain or modify it directly.

\begin{table}[H]
    \centering
    \small
    \caption{Summary of the literature audit. Algorithmic modification depth and implementation lineage use different denominators.}
    \label{tab:literature-audit-summary}
    \setlength{\arrayrulewidth}{0.35pt}
    \arrayrulecolor{black!50}
    \renewcommand{\arraystretch}{1.20}
    \begin{tabularx}{\textwidth}{|Y|r|r|}
        \hline
        \rowcolor{black!5}
        Audit quantity & Count & Share \\
        \hline
        Post-MeZO methods modifying 1 component & 11/30 & 36.7\% \\
        \hline
        Post-MeZO methods modifying 2 components & 14/30 & 46.7\% \\
        \hline
        Post-MeZO methods modifying 3 components & 5/30 & 16.7\% \\
        \hline
        Post-MeZO methods modifying all 4 components & 0/30 & 0.0\% \\
        \hline
        Methods modifying at most 2 components & 25/30 & 83.3\% \\
        \hline
        Methods with public code in direct or transitive MeZO lineage & 18/22 & 81.8\% \\
        \hline
        Methods with public code directly retaining/modifying MeZO code & 16/22 & 72.7\% \\
        \hline
    \end{tabularx}
\end{table}

Table~\ref{tab:four-component-taxonomy} gives representative examples of the four-component decomposition. The \emph{Execution effect} column summarizes how each method affects execution.

\begin{table}[H]
    \centering
    \scriptsize
    \caption{Representative LLM ZO methods under the four-component decomposition. Direction, Query, Estimator, and Update are the four algorithmic components; the final column summarizes the effect on execution.}
    \label{tab:four-component-taxonomy}
    \setlength{\tabcolsep}{4pt}
    \setlength{\arrayrulewidth}{0.35pt}
    \arrayrulecolor{black!50}
    \renewcommand{\arraystretch}{1.20}
    \begin{tabularx}{\textwidth}{|>{\raggedright\arraybackslash}p{1.85cm}|Y|Y|Y|Y|Y|}
        \hline
        \rowcolor{black!5}
        Method & Direction & Query & Estimator & Update & Execution effect \\
        \hline
        \mezo{}~\citep{malladi2023mezo}
        & Dense Gaussian direction; seed retained to regenerate it
        & Symmetric antithetic pair at $\pm\epsilon d$
        & Two-point finite-difference estimate
        & SGD-like parameter update; no method-specific optimizer state
        & Dense perturb / flip / restore of the model weights \\
        \hline
        \lozo{}~\citep{chen2025lozo}
        & $d=UV^T$ with right factor $V$ retained across multiple steps
        & Symmetric antithetic pair at $\pm\epsilon UV^T$
        & Projected two-point estimate
        & Base LoZO uses an SGD-like optimizer rule; optional optimizer variants are separate
        & Holding $V$ fixed keeps the queried perturbations and accumulated updates in low-rank factors \\
        \hline
        SparseMeZO~\citep{liu2025sparsemezo}
        & Selected parameter support / sparse direction
        & Antithetic pair on the selected support
        & Finite-difference estimate on that support
        & No separate optimizer-rule change; the parameter change follows the selected support
        & Reduces the parameters touched by each function evaluation and parameter change \\
        \hline
        HiZOO~\citep{zhao2024hizoo}
        & Hessian-informed perturbation geometry and curvature information
        & Center evaluation plus a symmetric perturbed pair
        & Two-point finite-difference estimate using the chosen geometry
        & Standard parameter update
        & Adds a center evaluation while keeping the perturbed evaluations forward-only \\
        \hline
        ZoAR~\citep{qiu2025zoar}
        & Gaussian perturbation
        & Fresh one-sided evaluations plus reuse of a bounded query buffer
        & Baseline averaging and aggregation of stored function values
        & Update rule can be chosen independently of reuse of earlier evaluations
        & Requires storing function values from earlier evaluations \\
        \hline
        ZO-Muon~\citep{lang2026zomuon}
        & Reduced-coordinate directions and persistent projection information
        & Multiple subspace finite-difference evaluations
        & Spectral orthogonalization of the estimated update
        & Applies the transformed update in the original parameter space
        & Exposes low-dimensional perturbations and a spectral transform \\
        \hline
        Adapter-space ZO~\citep{malladi2023mezo}
        & Direction in adapter parameter space
        & Usually one- or two-sided evaluations
        & Standard ZO estimate in reduced coordinates
        & Updates adapter parameters while base weights remain frozen
        & The optimized parameters are already separate from the base weights \\
        \hline
        ES at Scale~\citep{qiu2026esatscale}
        & Dense Gaussian full-parameter directions regenerated from seeds
        & Population of function evaluations
        & Reward normalization and reward-weighted aggregation
        & Additive ES update
        & Generation and reward evaluation replace two-point loss evaluation \\
        \hline
    \end{tabularx}
\end{table}

We assign each structural choice to the component that introduces it. Sparse support and low-rank geometry belong to Direction even though the resulting parameter change inherits that structure. Seeds, masks, subspaces, and other persistent information used to construct future perturbations also belong to Direction. Update records the optimizer rule and its state, including momentum, Adam moments, and stochastic variance-reduced gradient (SVRG) snapshots.

\section{Implementation Mapping and Backend Capabilities}\label{app:implementation-scope}
The four paper components map directly to implementation objects. Direction maps to \texttt{DirectionProvider}; Query maps to \texttt{ProbePlan}; Estimator maps to \texttt{ZOEstimator}; and Update maps to \texttt{UpdateTransform} producing an \texttt{UpdateIntent}. \texttt{ProbeObservation} stores the returned function values between \texttt{ProbePlan} execution and \texttt{ZOEstimator}.

\begin{table}[H]
    \centering\small
    \caption{Mapping from the four paper components to the implementation objects.}
    \label{tab:component-code-map}
    \setlength{\arrayrulewidth}{0.35pt}
    \arrayrulecolor{black!50}
    \renewcommand{\arraystretch}{1.20}
    \begin{tabularx}{\textwidth}{|p{.26\linewidth}|p{.28\linewidth}|Y|}
        \hline
        \rowcolor{black!5}
        Paper component & Main implementation object & Role \\
        \hline
        Direction & \texttt{DirectionProvider} & Constructs perturbations and stores information needed for future directions \\
        \hline
        Query & \texttt{ProbePlan} & Packages function-evaluation requests and their dependencies independently of the backend \\
        \hline
        Estimator & \texttt{ZOEstimator} & Uses returned function values to form a gradient estimate, including normalization, projection, or preconditioning \\
        \hline
        Update & \texttt{UpdateTransform} $\rightarrow$ \texttt{UpdateIntent} & Specifies the parameter update and optimizer state update \\
        \hline
    \end{tabularx}
\end{table}

The execution layer uses \texttt{ProbeExecutor}, \texttt{ZOStateBackend}, and the inference engine. \texttt{ProbeExecutor} consumes a \texttt{ProbePlan}, asks \texttt{ZOStateBackend} for the requested parameter state, and returns \texttt{ProbeObservation} values to the estimator. In the low-rank path evaluated in the main text, completed $(P^{(j)},V^{(j)})$ pairs map to LoRA adapter slots while the base weights remain unchanged during function evaluation. The PyTorch scoring backend uses the same low-rank factors, which enables the packing control used in the evaluation.

\section{Experimental Scope}\label{app:experimental-scope}

The main-text comparisons draw from several experiment families. The three-task learning study uses OPT-1.3B, rank 2, $\nu=50$, 1,000 steps, and seeds 42/43/44 on SST-2, BoolQ, and CB. The long-horizon study uses OPT-1.3B on SST-2 for 20k steps with the same three seeds, rank 2, batch size 16, and $\nu=50$. The matched-inference ladder uses SST-2, batch size 16, rank 2, $\nu=50$, sequence length 128, and the 15-model Qwen3/Llama/OPT suite from 350M to 32B. The co-serving experiments use Qwen3-8B: the numerical-isolation check records mixed model-runner batches, while the LoRA-prefill background comparison uses serving rates 10/20/30 requests/s and four windows per condition and rate.

\begin{table}[H]
    \centering\small
    \caption{Execution controls and their roles.}
    \label{tab:control-roles}
    \setlength{\arrayrulewidth}{0.35pt}
    \arrayrulecolor{black!50}
    \renewcommand{\arraystretch}{1.20}
    \begin{tabularx}{\linewidth}{|p{.34\linewidth}|Y|}
        \hline
        \rowcolor{black!5}
        Baseline or control & Purpose \\
        \hline
        Released LoZO & End-to-end speedup over the released implementation. \\
        \hline
        Low-rank and deferred-merge reference variants & Isolate changes in perturbation and update writes within the same reference backend. \\
        \hline
        Three-task LoZO learning comparison & Compare learning behavior across SST-2, BoolQ, and CB over three paired seeds. \\
        \hline
        Matched inference & Measure incremental ZO training cost after removing direction generation, perturbation preparation/application, and the final update while keeping the same request path. \\
        \hline
        15-model matched-inference ladder & Test whether the cost relationship persists across model families and scales. \\
        \hline
        Padded/packed PyTorch and packed vLLM & Separate input-padding effects from backend execution. \\
        \hline
        LoRA inference background under serving load & Compare shared GPU execution with ZO training under the same foreground load. \\
        \hline
    \end{tabularx}
\end{table}

\subsection{Hardware, Software, and Provenance}
\label{app:environment}

Table~\ref{tab:environment-provenance} summarizes the hardware, software, and execution settings used to interpret the main controls.

\begin{table}[H]
    \centering\footnotesize
    \caption{Recorded environment and execution configuration for the main experiment families.}
    \label{tab:environment-provenance}
    \setlength{\arrayrulewidth}{0.35pt}
    \arrayrulecolor{black!50}
    \renewcommand{\arraystretch}{1.20}
    \begin{tabularx}{\textwidth}{|p{.30\textwidth}|Y|}
        \hline
        \rowcolor{black!5}
        Study & Recorded environment / configuration \\
        \hline
        OPT cost, traffic, and packing controls & Per-run inputs, ranks, timing definitions, source revision, and result hashes; CUDA-graph control uses FP16 \\
        \hline
        15-model matched-inference ladder & RTX PRO 6000 Blackwell Server Edition host; checkpoint-native dtype; one engine reused within each model \\
        \hline
        Countdown local reference & One RTX PRO 6000 Blackwell Server Edition GPU; BF16; Python 3.12.13; one vLLM engine \\
        \hline
        Countdown dense-perturbation backend & BF16; greedy decoding; max length 2,048; max generated tokens 1,024; updates from an unperturbed weight copy \\
        \hline
        Co-serving background comparison & One RTX PRO 6000 Blackwell (96~GB); BF16; priority scheduling; prefix caching disabled; GPU-memory fraction 0.82 \\
        \hline
            20k larger-model and second-task learning & Qwen3-8B/SST-2 and Qwen3-0.6B/BoolQ; seed 42; rank 2; $\nu=50$ \\
        \hline
    \end{tabularx}
\end{table}

\section{Execution Correctness and Learning Correspondence}\label{app:correctness}

\paragraph{Strict state transitions.}
On OPT-1.3B, the 60-step SST-2 check gives 60/60 projected-gradient sign matches, projected-gradient Pearson correlation 0.999624, and $L_+/L_-$ correlations 0.999912/0.999895. BoolQ gives 58/60 sign matches and CB 60/60; all three runs have zero seed and refresh mismatches. On OPT-13B SST-2, the same 60-step protocol gives 59/60 sign matches, projected-gradient Pearson 0.99896, $L_+/L_-$ correlations 0.99977/0.99976, and zero seed and refresh mismatches. Over 1,000 OPT-13B steps, sign agreement is 960/1000 and projected-gradient Pearson is 0.98913; the maximum development-loss gap is 0.0049, or 1.13\% of the official loss drop, with final development accuracy 0.782 for released LoZO and 0.772 for \system{}.

\begin{table}[H]
    \centering
    \footnotesize
    \caption{Strict numerical correspondence. Values are measured directly between released LoZO and \system{}.}
    \label{tab:appendix-phase2}
    \begin{tabular}{lrrrr}
        \toprule
        Setting & Sign match & Pearson $c_t$ & Pearson $L_+$ & Pearson $L_-$ \\
        \midrule
        OPT-1.3B, SST-2, 60 steps & 60/60 & 0.999624 & 0.999912 & 0.999895 \\
        OPT-13B, SST-2, 60 steps & 59/60 & 0.99896 & 0.99977 & 0.99976 \\
        OPT-13B, SST-2, 1,000 steps & 960/1000 & 0.98913 & 0.99781 & 0.99774 \\
        \bottomrule
    \end{tabular}
\end{table}

\paragraph{Replicated short-horizon learning.}
The 1,000-step comparison uses three seeds on SST-2, BoolQ, and CB. Pooling the 15 seed-level development-loss checkpoints within each task gives Pearson correlations of 0.9923, 0.9957, and 0.9760, respectively. Pooling all 45 seed-level checkpoints gives Pearson correlation 0.9946 and linear-fit $R^2=0.9893$. Table~\ref{tab:learning-alignment} reports the endpoints used in the main text.

\paragraph{Long-horizon learning.}
The 20k-step OPT-1.3B SST-2 study uses seeds 42/43/44. Final mean development accuracy is 91.07\% for released LoZO and 91.40\% for \system{}; mean validation accuracy is 91.858\% and 92.278\%, respectively. The largest per-seed development-accuracy gap across the recorded checkpoints is 0.024 at step 4,000, and the final-step gap is 0.020. The corresponding released-LoZO seed ranges are 0.046 and 0.008. The largest block-averaged training-loss gap is 0.0377.

\section{Generation-Based Objectives and ES Results}
\label{app:es-evidence}
\subsection{Function Evaluations for Different Objectives}\label{app:query-shapes}
The required function evaluations depend on when the objective can be computed, rather than on the optimizer family.

\begin{table}[H]
    \centering
    \footnotesize
    \caption{Representative function evaluations for different objectives. The objective determines whether evaluation requires token scoring or autoregressive decoding.}
    \label{tab:posttraining-shapes}
    \setlength{\arrayrulewidth}{0.35pt}
    \arrayrulecolor{black!50}
    \renewcommand{\arraystretch}{1.20}
    \begin{tabularx}{\textwidth}{|>{\raggedright\arraybackslash}p{2.35cm}|Y|Y|Y|}
        \hline
        \rowcolor{black!5}
        Workload & When the objective is available & Function evaluation & Optimizer work after evaluation \\
        \hline
        ZO with a token-scoring objective (e.g., supervised NLL) & Available from provided/teacher-forced tokens & Token scoring / prefill & ZO estimator + parameter update; no backpropagation \\
        \hline
        First-order RL post-training~\citep{trl2026vllm,sheng2025hybridflow,hu2024openrlhf} & Depends on generated trajectories and rewards & Autoregressive rollout & Actor loss, backpropagation, optimizer step, trainer--rollout weight synchronization \\
        \hline
        Reward-based ZO (ES at Scale representative)~\citep{qiu2026esatscale} & Available only after generated responses are evaluated & Autoregressive generation under perturbed parameter values & Scalar reward aggregation + derivative-free update; no model backpropagation \\
        \hline
    \end{tabularx}
\end{table}

\subsection{ES on Countdown}
We use the ES-at-Scale Countdown task with Qwen/Qwen2.5-1.5B-Instruct and the official data and evaluation rule~\citep{qiu2026esatscale}. Each iteration uses 200 training rows, population 30, $\sigma=0.001$, learning rate $\alpha=0.0005$, z-score reward shaping, greedy decoding with up to 1,024 generated tokens, and one rollout per prompt. Evaluation uses the full 2,000-row split every five iterations for 500 iterations. Our dense-perturbation backend retains an unperturbed copy of the model weights in the worker and uses it to form dense Gaussian perturbations. It aggregates the population update in FP32 and writes the updated weights once per iteration. The batch-invariant run resumes from its checkpoint at step 200, testing checkpoint saving and resumption through the same loop.

\begin{table}[H]
    \centering
    \footnotesize
    \caption{Countdown accuracy for local runs under the ES-at-Scale workload configuration.}
    \label{tab:es-countdown}
    \begin{tabular}{@{}lcc@{}}
        \toprule
        Run & Final accuracy & Best accuracy \\
        \midrule
        Authors' vLLM repository, local run & 28.45\% & 29.25\% (465) \\
        Our dense-perturbation backend & 29.20\% & 30.15\% (485) \\
        Our backend + batch-invariant mode & 31.20\% & 31.60\% (490) \\
        Our async DP2 backend, repetition penalty 1.1 & \textbf{35.25\%} & \textbf{36.15\% (485)} \\
        \bottomrule
    \end{tabular}
\end{table}

A separate three-update consistency run uses population 8 and seed 42. Initial weights match on 198/198 parameter tensors. At step 1, all eight perturbation/restore states match bit-for-bit. Applying our z-score shaping to the reference rewards matches 8/8 values at each of the three steps; replaying our update from the reference pre-update weights with its seeds and rewards changes 0 of 1,543,714,304 elements relative to the reference update at steps 1--3. Under end-to-end generation, the two vLLM versions produce 16/32 identical unperturbed greedy texts at step 0, and the subsequent reward sequences differ.

\section{Multivariate Central Limit Theorem}\label{app:clt-proof}
For one matrix, let
\begin{equation}
    Z_r=\frac{1}{\sqrt r}UV^T,\qquad U\in\mathbb R^{m\times r},\quad V\in\mathbb R^{n\times r},
\end{equation}
with independent standard-normal entries. For a fixed entry,
\begin{equation}
    (Z_r)_{ij}=\frac{1}{\sqrt r}\sum_{k=1}^{r}u_{ik}v_{jk}.
\end{equation}
Each summand has mean zero and variance one, so the scalar central limit theorem gives
\begin{equation}
    (Z_r)_{ij}
    =\frac{1}{\sqrt r}\sum_{k=1}^{r}u_{ik}v_{jk}
    \xrightarrow{d}\mathcal N(0,1).
    \label{eq:clt}
\end{equation}
For any fixed finite set of distinct entries $(i_1,j_1),\ldots,(i_s,j_s)$, the corresponding vector is a normalized sum of i.i.d. zero-mean random vectors across $k$. Distinct entries have zero covariance because at least one independent zero-mean $U$ or $V$ factor differs, so the covariance matrix is the identity. The multivariate central limit theorem therefore gives
\begin{equation}
    \big((Z_r)_{i_1j_1},\ldots,(Z_r)_{i_sj_s}\big)
    \xrightarrow{d}\mathcal N(0,I_s).
\end{equation}

\section{Finite-Rank Directions and Update-Rule Decoupling}\label{app:rank-continuum}

The 20k-step appendix sweep uses Qwen3-0.6B, Llama-3.2-1B, and OPT-1.3B on SST-2, with three seeds per arm and one learning rate per model/update rule shared by dense and low-rank Directions. Table~\ref{tab:appendix-phase6-curves} reports step-20k development loss for the Gaussian-product Direction family; the entries are low-rank minus dense, paired by seed.

\begin{table}[H]
    \centering
    \footnotesize
    \caption{Step-20k development loss for finite-rank Gaussian-product Directions. Delta columns are low-rank minus dense, mean $\pm$ sample standard deviation over three paired seeds.}
    \label{tab:appendix-phase6-curves}
    \begin{tabular}{lrrrr}
        \toprule
        Model & Dense loss & $\Delta r=1$ & $\Delta r=16$ & $\Delta r=64$ \\
        \midrule
        Qwen3-0.6B & 0.3327 & $+0.0112\pm0.0115$ & $+0.0338\pm0.0236$ & $+0.0034\pm0.0074$ \\
        Llama-3.2-1B & 0.2235 & $+0.0025\pm0.0082$ & $-0.0023\pm0.0169$ & $-0.0031\pm0.0009$ \\
        OPT-1.3B & 0.2247 & $+0.0092\pm0.0170$ & $+0.0050\pm0.0050$ & $-0.0019\pm0.0187$ \\
        \bottomrule
    \end{tabular}
\end{table}

A separate low-load timing sweep spans ranks 1--512. At rank 1, step time is 33.24 vs. 39.42~ms on Qwen3-0.6B, 21.28 vs. 46.52~ms on Llama-3.2-1B, and 27.81 vs. 53.00~ms on OPT-1.3B, each relative to its dense Direction.

The same low-rank Direction is also combined with ZO-SGD-Sign and DPZero~\citep{zhang2024dpzero}. Table~\ref{tab:update-rule-grid} reports the paired step-20k development-loss difference from each rule's dense arm.

\begin{table}[H]
    \centering
    \footnotesize
    \caption{Estimator/update-rule decoupling at 20k steps. Entries are low-rank minus dense development loss, averaged over three paired seeds.}
    \label{tab:update-rule-grid}
    \begin{tabular}{lrrrrr}
        \toprule
        Model & Sign $\nu=1$ & Sign $\nu=50$ & DPZero $\nu=1$ & DPZero $\nu=50$ & ZO-SGD $\nu=50$ \\
        \midrule
        Qwen3-0.6B & +0.0160 & +0.0290 & +0.0018 & +0.0097 & +0.0204 \\
        Llama-3.2-1B & +0.0067 & +0.0071 & -0.0010 & -0.0078 & +0.0020 \\
        OPT-1.3B & +0.0098 & +0.0028 & -0.0067 & -0.0182 & +0.0027 \\
        \bottomrule
    \end{tabular}
\end{table}

\paragraph{Larger-model and second-task runs.}
We extend the 20k-step comparison to Qwen3-8B on SST-2 and Qwen3-0.6B on BoolQ, using seed 42, rank $r=2$, and $\nu=50$. Each of ZO-SGD, ZO-SGD-Sign, and DPZero is evaluated with dense Gaussian and low-rank directions. The runs use 1,000 training examples, batch size 16, perturbation scale $\epsilon=10^{-3}$, and constant learning rates. Development loss uses 500 examples; validation accuracy uses 872 SST-2 examples or 1,000 BoolQ examples. Table~\ref{tab:appendix-scale-task-extension} reports the final step-20k checkpoint.

One learning rate is shared by both directions within each model/task and rule. Rates are selected on a $1/2/5$ grid using seed 42, based on the final increase from the minimum development loss and from the minimum 2,000-step-averaged training loss, with a threshold of 0.02. Selection uses the dense arm for the three small-model SST-2 settings and the rank-2, $\nu=50$ arm for the two added settings. The selected rates for the added settings appear in the table.

\begin{table}[H]
    \centering
    \footnotesize
    \setlength{\tabcolsep}{4pt}
    \caption{Larger-model and second-task learning at step 20,000 (seed 42). Both directions use the listed learning rate; the low-rank arm uses $r=2$, $\nu=50$. Accuracy is a fraction on the validation split.}
    \label{tab:appendix-scale-task-extension}
    \begin{tabular}{lrrrrr}
        \toprule
        & & \multicolumn{2}{c}{Dev. loss} & \multicolumn{2}{c}{Valid. accuracy} \\
        \cmidrule(lr){3-4}\cmidrule(lr){5-6}
        Rule & Learning rate & Dense & Low-rank & Dense & Low-rank \\
        \midrule
        \multicolumn{6}{l}{\textbf{Qwen3-8B / SST-2}} \\
        ZO-SGD & $5\times10^{-7}$ & 0.2788 & 0.2892 & 0.9186 & 0.9140 \\
        ZO-SGD-Sign & $2\times10^{-5}$ & 0.2650 & 0.2733 & 0.9197 & 0.9232 \\
        DPZero & $2\times10^{-7}$ & 0.3750 & 0.4095 & 0.8612 & 0.8647 \\
        \midrule
        \multicolumn{6}{l}{\textbf{Qwen3-0.6B / BoolQ}} \\
        ZO-SGD & $2\times10^{-7}$ & 0.5867 & 0.5207 & 0.7230 & 0.7250 \\
        ZO-SGD-Sign & $1\times10^{-5}$ & 0.5490 & 0.5423 & 0.7240 & 0.7220 \\
        DPZero & $2\times10^{-7}$ & 0.6217 & 0.5631 & 0.7210 & 0.7140 \\
        \bottomrule
    \end{tabular}
\end{table}

\section{Timing Accounting}\label{app:timing-accounting}

For the clean Qwen3-14B batch-16 control, the forward and reversed matched-inference validation sessions average 92.9939~ms for \system{} and 91.9260~ms for matched inference, giving $1.162\%\pm0.400$ percentage points of overhead.

The 15-model sweep uses one diagnostic timing row per model. Table~\ref{tab:clean-dense-ladder} gives the values plotted in Figure~\ref{fig:matched-inference-scaling}(a), with matched-inference overhead computed by removing direction generation, perturbation preparation and application, and the final update from the measured ZO step. Time outside the measured model-evaluation region is 2.28--5.28~ms and follows a pooled diagnostic fit of 1.181~ms $+$ 9.45~$\mu$s per LoRA module ($R^2=0.925$).

\begin{table}[H]
    \centering
    \scriptsize
    \caption{15-model matched-inference overhead accounting (batch 16, rank 2, $\nu=50$). Overhead is $100(T_{\mathrm{ZO}}/T_{\mathrm{matched}}-1)$ with $T_{\mathrm{matched}}$ formed by removing direction generation, perturbation preparation/application, and the final update from the measured ZO step.}
    \label{tab:clean-dense-ladder}
    \begin{tabular}{lrrrr}
        \toprule
        Model & Params (B) & ZO (ms) & Matched inf. (ms) & Overhead (\%) \\
        \midrule
        Qwen3-0.6B & 0.6 & 26.13 & 24.12 & +8.320 \\
        Qwen3-1.7B & 1.7 & 28.23 & 26.05 & +8.393 \\
        Qwen3-4B & 4 & 36.94 & 34.56 & +6.891 \\
        Qwen3-8B & 8 & 57.01 & 54.35 & +4.900 \\
        Qwen3-14B & 14 & 92.22 & 89.25 & +3.324 \\
        Qwen3-32B & 32 & 204.18 & 199.79 & +2.197 \\
        \midrule
        Llama-3.2-1B & 1.2 & 17.70 & 16.06 & +10.226 \\
        Llama-3.2-3B & 3.2 & 30.65 & 28.39 & +7.958 \\
        Llama-3.1-8B & 8 & 57.24 & 54.50 & +5.026 \\
        \midrule
        OPT-350M & 0.35 & 25.06 & 23.13 & +8.349 \\
        OPT-1.3B & 1.3 & 24.53 & 22.67 & +8.176 \\
        OPT-2.7B & 2.7 & 30.26 & 28.06 & +7.848 \\
        OPT-6.7B & 6.7 & 52.89 & 50.55 & +4.615 \\
        OPT-13B & 13 & 92.42 & 89.24 & +3.572 \\
        OPT-30B & 30 & 200.98 & 197.76 & +1.628 \\
        \bottomrule
    \end{tabular}
\end{table}

\section{Throughput Against the Released LoZO Implementation}
\label{app:released-lozo}

The released \lozo{} reference implementation is evaluated under the same batch-16 timing protocol as \system{}. Table~\ref{tab:appendix-released-lozo-scaling} reports the recorded tail-100 mean end-to-end step time at batch 16 for 15 paper models. The resulting speedups range from 2.10$\times$ to 9.94$\times$.

\begin{table}[H]
    \centering
    \footnotesize
    \caption{End-to-end step time against released \lozo{} at batch 16. Each row reports the recorded tail-100 mean for that paired run.}
    \label{tab:appendix-released-lozo-scaling}
    \begin{tabular}{lrrr}
        \toprule
        Model & \lozo{} step (s) & \system{} step (s) & Speedup \\
        \midrule
        Qwen3-0.6B & 0.0826 & 0.0263 & 3.14$\times$ \\
        Qwen3-1.7B & 0.0968 & 0.0270 & 3.58$\times$ \\
        Qwen3-4B & 0.1971 & 0.0370 & 5.33$\times$ \\
        Qwen3-8B & 0.4910 & 0.0602 & 8.15$\times$ \\
        Qwen3-14B & 0.9110 & 0.0978 & 9.32$\times$ \\
        Qwen3-32B & 2.0750 & 0.2087 & 9.94$\times$ \\
        \midrule
        Llama-3.2-1B & 0.0674 & 0.0172 & 3.93$\times$ \\
        Llama-3.2-3B & 0.1592 & 0.0310 & 5.14$\times$ \\
        Llama-3.1-8B & 0.4910 & 0.0595 & 8.25$\times$ \\
        \midrule
        OPT-350M & 0.0515 & 0.0245 & 2.10$\times$ \\
        OPT-1.3B & 0.0648 & 0.0242 & 2.68$\times$ \\
        OPT-2.7B & 0.1211 & 0.0295 & 4.11$\times$ \\
        OPT-6.7B & 0.3717 & 0.0561 & 6.62$\times$ \\
        OPT-13B & 0.7339 & 0.0957 & 7.67$\times$ \\
        OPT-30B & 1.8764 & 0.2035 & 9.22$\times$ \\
        \bottomrule
    \end{tabular}
\end{table}

\section{Auxiliary Runtime Controls}\label{app:runtime-controls}
CUDA graphs provide a small fixed benefit in the tested scoring workload: 2.1\% and 0.9\% speedups at 512 and 2048 tokens, respectively.

\section{Correctness with Mixed Request Batches}\label{app:mixed-batch}

\begin{table}[H]
    \centering
    \footnotesize
    \caption{Numerical isolation when one scheduled Qwen3-8B batch contains both foreground serving requests and background ZO probes.}
    \label{tab:mixed-correctness}
    \begin{tabular}{lr}
        \toprule
        Check & Result \\
        \midrule
        Physical mixed batches & 31 \\
        Batch sizes exercised & 4, 8, 16 \\
        Foreground requests checked & 14 \\
        Paired ZO steps checked & 15 \\
        Maximum numerical difference & 0 \\
        Base weights unchanged & yes \\
        \bottomrule
    \end{tabular}
\end{table}

\section{LoRA Inference Background under Serving Load}
\label{app:resource-evidence}

The shared-load experiment holds foreground traffic fixed on one long-lived Qwen3-8B engine. It compares a LoRA-prefill background with the same background plus ZO updates under two implementation variants (fix 5 and fixes 4+5). The replay uses identical prompt token IDs, the same ordering barrier between the positive/negative evaluation pair, perturbation rank 8, and background priority 1 (serving priority 0). ZO uses BoolQ prompts, batch size 16, allocated LoRA rank 128, $\nu=50$, and learning rate $10^{-7}$. Serving requests use 256-token prompts, 128-token outputs, burstiness 0.3, and rates 10/20/30 requests/s. Each rate has four measurements per condition in mirrored order.

Table~\ref{tab:resource-absolute} reports sustained-load means. Relative to the corresponding LoRA-prefill background, the ZO arms retain 98.6--99.1\% of foreground request throughput and 97.2--97.9\% of background scheduled-token throughput across the tested rates.

\begin{table}[H]
    \centering
    \scriptsize
    \caption{Absolute shared-load measurements for the co-serving comparison ($n=4$ per condition and foreground request rate).}
    \label{tab:resource-absolute}
    \setlength{\tabcolsep}{3.5pt}
    \begin{tabular}{@{}llrrrr@{}}
        \toprule
        FG RPS & Condition & FG req/s & BG sched. tok/s & Total sched. tok/s & Batch fill \\
        \midrule
        10 & LoRA-prefill & 5.114 & 16,938 & 18,871 & 1.000 \\
        10 & ZO (fix 5) & 5.069 & 16,515 & 18,426 & 1.000 \\
        10 & ZO (fixes 4+5) & 5.054 & 16,458 & 18,374 & 1.000 \\
        \midrule
        20 & LoRA-prefill & 10.178 & 13,666 & 17,534 & 0.885 \\
        20 & ZO (fix 5) & 10.053 & 13,369 & 17,193 & 0.882 \\
        20 & ZO (fixes 4+5) & 10.076 & 13,343 & 17,172 & 0.878 \\
        \midrule
        30 & LoRA-prefill & 14.823 & 11,273 & 16,904 & 0.771 \\
        30 & ZO (fix 5) & 14.638 & 11,019 & 16,574 & 0.770 \\
        30 & ZO (fixes 4+5) & 14.610 & 11,039 & 16,581 & 0.769 \\
        \bottomrule
    \end{tabular}
\end{table}

Figure~\ref{fig:coserving-throughput} normalizes the two ZO arms to the LoRA-prefill background, so the remaining gap directly measures the cost of inserting the ZO update path into the shared engine. SLO-aware admission can be supplied by online/offline serving schedulers such as HyGen~\citep{sun2025hygen}.

\section{LoRA Bank Capacity and Cyclic Reuse}\label{app:bank-runtime}

The configured bank rank trades fold frequency against fixed adapter cost. Table~\ref{tab:bank-runtime} reports the full OPT-2.7B step time at rank 2, $\nu=10$. Final accuracy stays between 0.730 and 0.732 across the sweep.

\begin{table}[H]
    \centering
    \small
    \caption{Bank capacity, fold frequency, and end-to-end step time on OPT-2.7B.}
    \label{tab:bank-runtime}
    \begin{tabular}{rrrr}
        \toprule
        Bank rank & Folds & Tail-100 step (ms) & Final accuracy \\
        \midrule
        2 & 1024 & 49.20 & 0.732 \\
        4 & 102 & 41.47 & 0.732 \\
        8 & 34 & 38.30 & 0.732 \\
        16 & 14 & 37.24 & 0.730 \\
        32 & 6 & 39.32 & 0.732 \\
        64 & 3 & 37.03 & 0.732 \\
        128 & 1 & 38.02 & 0.730 \\
        256 & 0 & 42.21 & 0.732 \\
        512 & 0 & 53.54 & 0.732 \\
        \bottomrule
    \end{tabular}
\end{table}

Cyclic subspace reuse bounds the bank capacity while keeping the base weights read-only. In 20k-step SST-2 runs on Qwen3-0.6B, Llama-3.2-1B, and OPT-1.3B, bank-64 and bank-16 wrap variants complete three seeds each with zero base folds and unchanged base-state digests in all 18 runs. Relative to the direct-update arm, the mean step-20k development-loss differences for bank-64/bank-16 wrapping are +0.0240/+0.0194 on Qwen3-0.6B, -0.0111/+0.0015 on Llama-3.2-1B, and +0.0054/-0.0007 on OPT-1.3B.

\paragraph{Larger-model and second-task bank runs.}
The same rank-64 bank is also evaluated on Qwen3-8B SST-2 and Qwen3-0.6B BoolQ for 20k steps with seed 42, rank $r=2$, and $\nu=50$. Table~\ref{tab:bank-scale-task-extension} compares cyclic reuse and merging on overflow with the corresponding direct-update arm. Each added cyclic-reuse run completes 12 wraps without a base merge. Together with the three-model, three-seed study, all 20 cyclic-reuse runs pass the full base-state digest check.

\begin{table}[H]
    \centering
    \footnotesize
    \setlength{\tabcolsep}{4pt}
    \caption{Rank-64 bank policies on the larger-model and second-task runs (ZO-SGD, seed 42, step 20,000). Loss differences are relative to the rank-2 direct-update arm in the same setting. Accuracy is a validation fraction; dashes denote quantities not applicable to direct update.}
    \label{tab:bank-scale-task-extension}
    \begin{tabular}{lrrrrr}
        \toprule
        Policy & Dev. loss & Valid. acc. & $\Delta$ dev. loss & Wraps & Base merges \\
        \midrule
        \multicolumn{6}{l}{\textbf{Qwen3-8B / SST-2}} \\
        Direct update & 0.2892 & 0.9140 & -- & -- & -- \\
        Cyclic reuse & 0.2602 & 0.9255 & -0.0290 & 12 & 0 \\
        Merge on overflow & 0.2679 & 0.9266 & -0.0213 & 0 & 12 \\
        \midrule
        \multicolumn{6}{l}{\textbf{Qwen3-0.6B / BoolQ}} \\
        Direct update & 0.5207 & 0.7250 & -- & -- & -- \\
        Cyclic reuse & 0.5184 & 0.7390 & -0.0024 & 12 & 0 \\
        Merge on overflow & 0.5272 & 0.7360 & +0.0064 & 0 & 12 \\
        \bottomrule
    \end{tabular}
\end{table}

\section{Quantized Base-Weight Compatibility}\label{app:quantized-base}
The frozen base can remain quantized throughout \system{} training. On Qwen3-8B SST-2, both BF16 and FP8 runs preserve an identical full-state digest before and after training. FP8 reduces the measured step time from 64.24 to 47.22~ms (1.36$\times$) while reaching similar 1,000-step learning endpoints.

\begin{table}[H]
    \centering
    \footnotesize
    \caption{Qwen3-8B with frozen base weights and FP8. Learning values are at step 1,000; timing uses the final 100 steps of the 200-step timing run.}
    \label{tab:appendix-quantized-base}
    \begin{tabular}{lrrrr}
        \toprule
        Base & Step (ms) & Final dev. loss & Final dev. acc. & Final valid acc. \\
        \midrule
        BF16 & 64.24 & 0.4904 & 0.782 & 0.780 \\
        FP8  & 47.22 & 0.4933 & 0.770 & 0.784 \\
        \bottomrule
    \end{tabular}
\end{table}

\end{document}